%% file: main.tex
\documentclass{article}

\usepackage[preprint, nonatbib]{neurips_2025}

\usepackage[utf8]{inputenc} %
\usepackage[T1]{fontenc}    %
\usepackage{booktabs}       %
\usepackage{amsfonts}       %
\usepackage{nicefrac}       %
\usepackage{microtype}      %
\usepackage{xcolor}         %
\input{pkg}

\input{math}

\usepackage{url}
\usepackage[pagebackref=true,breaklinks=true,colorlinks,bookmarks=false]{hyperref}
\hypersetup{
    colorlinks=true,
    filecolor=magenta,      
    urlcolor=magenta,
}

\usepackage{xspace}
\newcommand{\benchmark}{{\color{black}{VCRBench}}\xspace}
\newcommand{\giticon}[1][1em]{\includegraphics[height=#1]{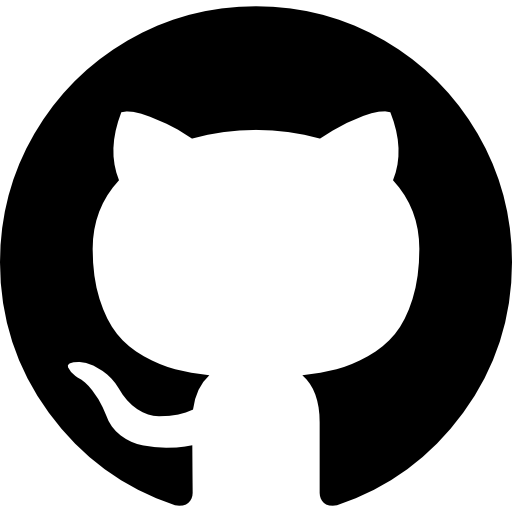}}
\newcommand{\hficon}[1][1em]{\includegraphics[height=#1]{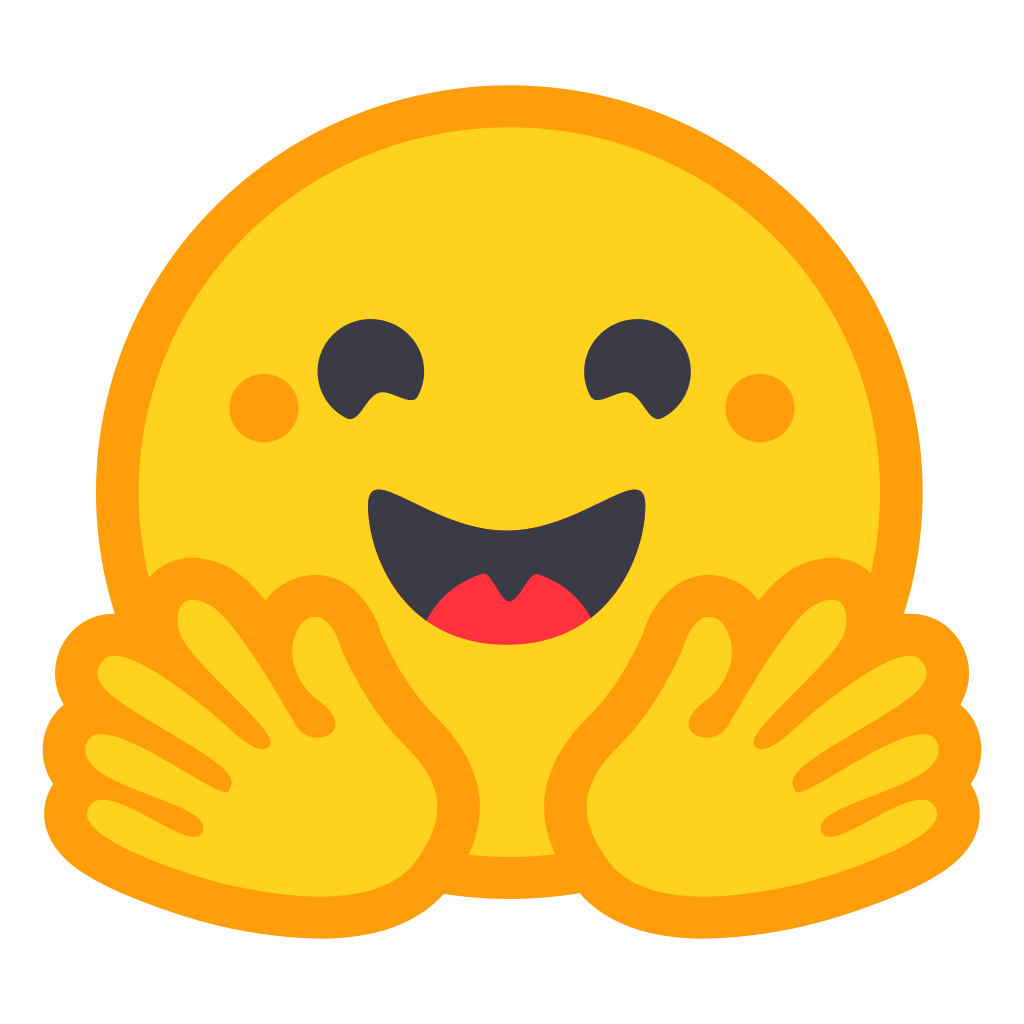}}
\newcommand{\webicon}[1][1em]{\includegraphics[height=#1]{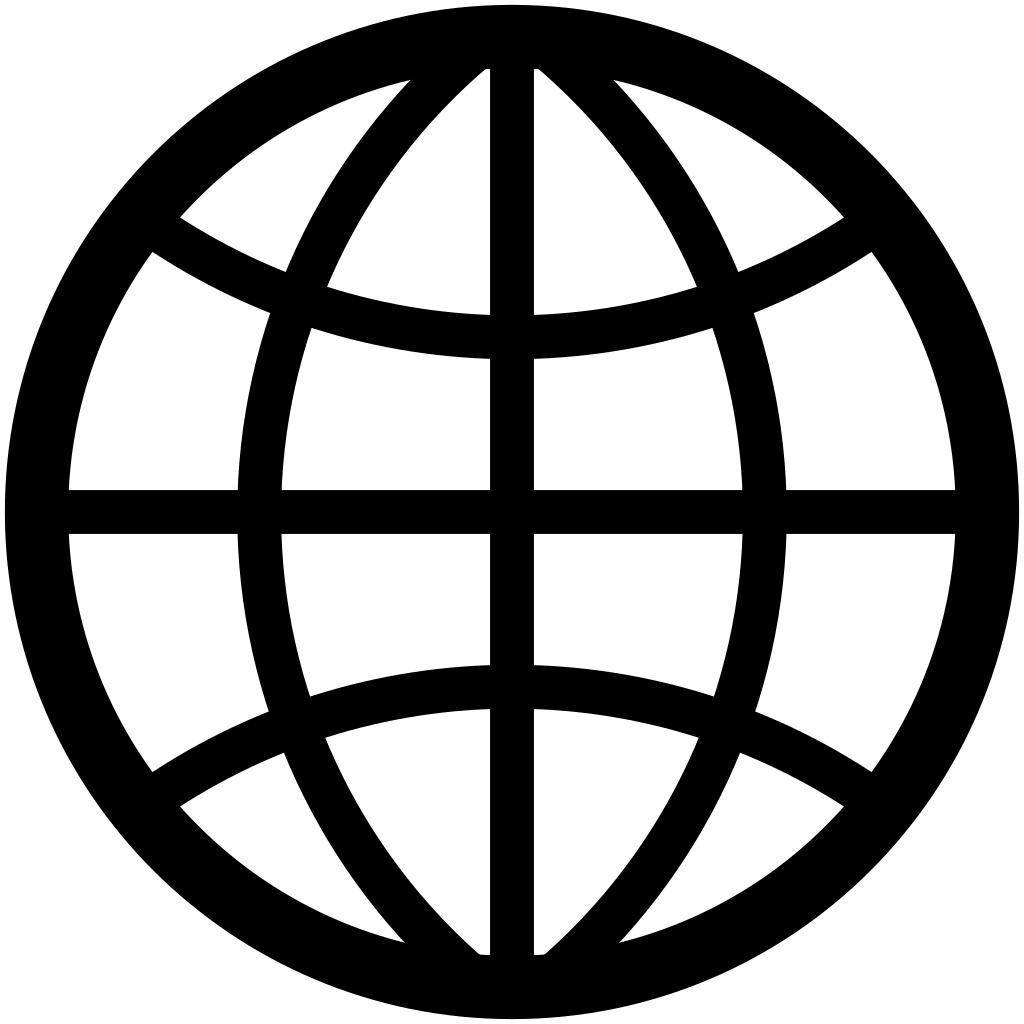}}

\title{
VCRBench: Exploring Long-form Causal Reasoning Capabilities of Large Video Language Models
}

\author{%
  Pritam Sarkar \\
  Queen's University, Canada and Vector Institute\\
  \texttt{pritam.sarkar@queensu.ca} \\
  \And
  Ali Etemad \\
  Queen's University, Canada \\
  \texttt{ali.etemad@queensu.ca} \\
\AND
{\begin{tabular}{cccc}
\href{https://pritamsarkar.com/VCRBench}{\webicon[16pt]}
& \href{https://github.com/pritamqu/VCRBench}{\giticon[16pt]}
& \href{https://huggingface.co/datasets/pritamqu/VCRBench}{\hficon[16pt]}
\\
Website & Code & Data \\
\end{tabular}}
}

\begin{document}

\maketitle

\setcounter{footnote}{0}

\input{main_paper}

\section*{Acknowledgment}
We thank Debaditya Shome and Nishq Poorav Desai for their help in building the platform used to collect responses from human evaluators on \benchmark.
We also thank members of our lab who volunteered as human evaluators in assessing \benchmark.
We thank the Bank of Montreal and Mitacs for funding this research, and the Vector Institute for providing computational resources.

{
\bibliographystyle{unsrt}
\bibliography{egbib}
}

\clearpage
\appendix

\begin{center}
    \maketitle
    \Large
    \textbf{Appendix}
\end{center}

\input{appx}

\end{document}

%% file: pkg.tex
\usepackage{colortbl}
\usepackage{multirow}
\definecolor{fade}{gray}{0.4}
\definecolor{green}{rgb}{0.0, 0.5, 0.0}
\definecolor{grey}{rgb}{0.5, 0.5, 0.5}

\newcommand\fadetext[1]{\textcolor{grey}{#1}}
\newcommand{\inc}[1]{\textcolor{green}{\footnotesize$_\mathbf{{\uparrow#1}}$}}

\newcommand{\decbad}[1]{\textcolor{red}{\footnotesize$_\mathbf{{\downarrow#1}}$}}

\usepackage{soul}
\newcommand{\specialcell}[2][c]{%
\begin{tabular}[#1]{@{}c@{}}#2\end{tabular}}
\newcommand{\specialcellleft}[2][l]{%
\begin{tabular}[#1]{@{}l@{}}#2\end{tabular}}

\usepackage{fontawesome}

\usepackage{pifont}%
\usepackage{graphicx}
\usepackage{amsmath}
\usepackage{booktabs}
\usepackage{enumitem}
\usepackage{color, colortbl}
\usepackage{multirow}
\usepackage{diagbox}
\usepackage{algorithm}
\usepackage{listings}
\definecolor{rc1}{RGB}{235,235,235}
\definecolor{rc2}{RGB}{255,255,255}
\definecolor{codeblue}{rgb}{0.25,0.5,0.5}
\definecolor{codekw}{rgb}{0.85, 0.18, 0.50}
\usepackage{booktabs}       %
\usepackage{amsfonts}       %
\usepackage{nicefrac}       %
\usepackage{microtype}      %
\usepackage{graphicx}
\usepackage{tikz}
\usepackage{comment}
\usepackage{amsmath,amssymb} %
\usepackage{color}

\usepackage{stackengine}

\usepackage{algpseudocode}

\newcommand\blue[1]{{\color{blue}{#1}}}
\newcommand\red[1]{{\color{red}{#1}}}

\definecolor{green}{rgb}{0.0, 0.5, 0.0}

\definecolor{darkgreen}{rgb}{0.0, 0.4, 0.0}
\newcommand\darkgreen[1]{{\color{darkgreen}{#1}}}

\usepackage{caption}
\usepackage{subcaption}

\usepackage[most]{tcolorbox}
\usepackage{mdframed}
\usepackage{wrapfig}

\usepackage{bbm}
\usepackage{fix-cm} %

\usepackage{xtab}

\usepackage{alltt,xcolor}
\usepackage{relsize}
\usepackage{rotating} %
\usepackage{amsmath,amssymb,amsthm}

\newcounter{qcounter}

\usepackage{verbatim}

\newcommand{\goldicon}[1][1em]{\includegraphics[height=#1]{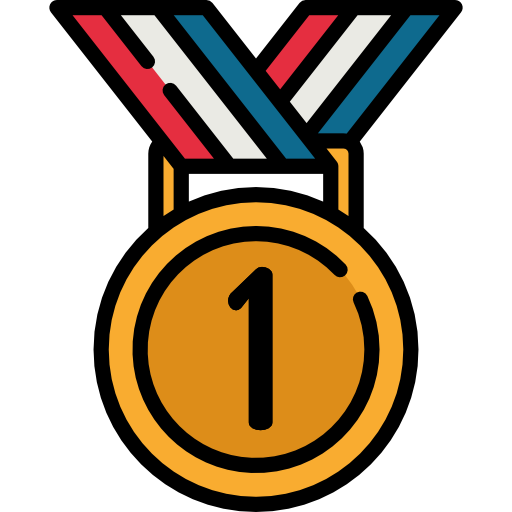}}
\newcommand{\silvericon}[1][1em]{\includegraphics[height=#1]{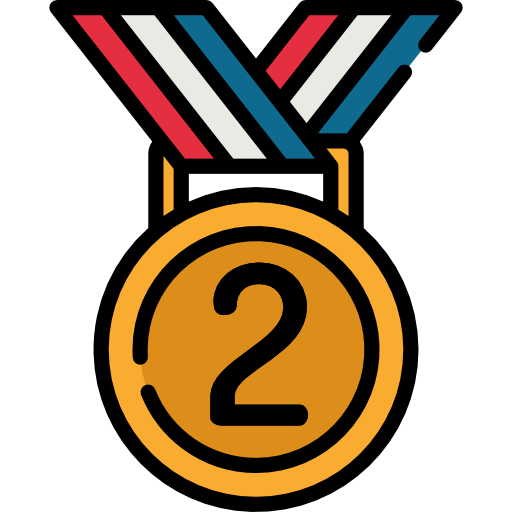}}
\newcommand{\bronzeicon}[1][1em]{\includegraphics[height=#1]{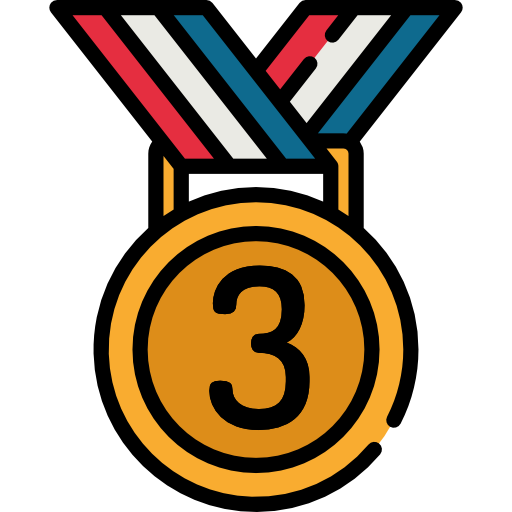}}

\newtcblisting[auto counter,number within=chapter]{sourcecode}[2][]{
  sharp corners,
  breakable,
  fonttitle=\bfseries,
  colframe=black!70,
  listing only,
  listing options={
    basicstyle=\ttfamily\footnotesize,
    language=php,
    showstringspaces=false,
    breakatwhitespace=true,
    breaklines=true,
    postbreak=\mbox{},
    tabsize=2,
    linewidth=\linewidth,
    breakindent=0pt
  },
  title=#2,
  #1
}

\newenvironment{prompt}[1][\textwidth] %
{
    \begin{tcolorbox}[
        width=#1, %
        colframe=black, %
        colback=gray!10, %
        sharp corners, %
        boxrule=0.5mm, %
        left=5pt, right=5pt, top=5pt, bottom=5pt, %
        fontupper=\ttfamily %
    ]
    \tiny
}
{
    \end{tcolorbox}
}

\newenvironment{prompt2}[1][\textwidth] %
{
    \begin{tcolorbox}[
        width=#1, %
        colframe=black, %
        colback=gray!10, %
        sharp corners, %
        boxrule=0.5mm, %
        left=5pt, right=5pt, top=5pt, bottom=5pt, %
        fontupper=\ttfamily %
    ]
    \small
}
{
    \end{tcolorbox}
}

{
    \begin{tcolorbox}[
        width=#1, %
        colframe=black, %
        colback=gray!10, %
        sharp corners, %
        boxrule=0.5mm, %
        left=5pt, right=5pt, top=5pt, bottom=5pt, %
        fontupper=\ttfamily %
    ]
    \tiny
}
{
    \end{tcolorbox}
}

%% file: math.tex
\usepackage{amsmath,amsfonts,bm}

\def\eqref#1{equation~\ref{#1}}

\def\1{\bm{1}}

\DeclareMathAlphabet{\mathsfit}{\encodingdefault}{\sfdefault}{m}{sl}
\SetMathAlphabet{\mathsfit}{bold}{\encodingdefault}{\sfdefault}{bx}{n}

\def\gD{{\mathcal{D}}}

%% file: main_paper.tex
\begin{abstract}
Despite recent advances in video understanding, the capabilities of Large Video Language Models (LVLMs) to perform video-based causal reasoning remains underexplored, largely due to the absence of relevant and dedicated benchmarks for evaluating \textit{causal reasoning} in visually grounded and goal-driven settings. To fill this gap, we introduce a novel benchmark named {Video-based long-form Causal Reasoning (\benchmark)}. We create \benchmark using procedural videos of simple everyday activities, where the steps are deliberately shuffled with each clip capturing a key causal event, to test whether LVLMs can identify, reason about, and correctly sequence the events needed to accomplish a specific goal. Moreover, the benchmark is carefully designed to prevent LVLMs from exploiting linguistic shortcuts, as seen in multiple-choice or binary QA formats, while also avoiding the challenges associated with evaluating open-ended QA. Our evaluation of state-of-the-art LVLMs on \benchmark suggests that these models struggle with video-based long-form causal reasoning, primarily due to their difficulty in modeling long-range causal dependencies directly from visual observations. As a simple step toward enabling such capabilities, we propose {Recognition-Reasoning Decomposition (RRD)}, a modular approach that breaks video-based causal reasoning into two sub-tasks of \textit{video recognition} and \textit{causal reasoning}. Our experiments on \benchmark show that RRD significantly boosts accuracy on \benchmark, with gains of up to $25.2\%$. 
Finally, our thorough analysis reveals interesting insights into the reasoning capabilities of LVLMs, for instance, that they primarily rely on their \textit{language} knowledge when tackling complex \textit{video}-based long-form causal reasoning tasks.

\end{abstract}

\section{Introduction}
\label{sec:intro}

Long-form causal reasoning in video involves structured and goal-directed analysis of sequences of visual events. Such capabilities are essential for real-world applications such as household and industrial robotics~\cite{sermanet2024robovqa,li2024mmro}, embodied AI agents \cite{egoplan,majumdar2024openeqa,chen2023towards}, spatial intelligence systems~\cite{feng2025survey,chen2024spatialvlm,huang2023voxposer}, and assistive technologies~\cite{liu2024spectrum,bai2024m3d}, all of which rely on reasoning about causally dependent visual events. While recent advances in vision-language modeling~\cite{clip,siglip,internvideo2} have led to the development of powerful large video language models (LVLMs)~\cite{longvu,slowfast_video,videochat2,llamavid,videollama2,videollama3,timechat,longvila,llavanextvid,videollava,gemini15,internvl,internvl25,qwenvl,qwenvl25}, their ability to perform long-form causal-reasoning based on visual observations remains largely underexplored. This is in part due to the lack of benchmarks specifically designed to evaluate causal reasoning in visually-grounded goal-driven settings.
In this work, we take a step toward filling this gap by systematically evaluating the video-based causal reasoning capabilities of state-of-the-art LVLMs through a new benchmark. 
Building on this, we also design a simple modular approach to enhance LVLM performance on video-based long-form causal reasoning tasks.

\begin{table}[]
\centering
\begin{minipage}{0.33\linewidth}
\centering
\texttt{\tiny <Question>}
\resizebox{!}{1.1\linewidth}{
\begin{prompt}%
The given video consists of multiple clips, each showing a different segment needed to complete the task: \blue{Make Lemonade}.

These clips are randomly shuffled, and your job is to arrange them in the correct order to complete the task: \blue{Make Lemonade}.

The clip numbers are mentioned at the beginning of each clip as Clip 1, Clip 2, and so on.
In order to solve this task, first, you should identify the activity that is performed in each clip, and then use your reasoning and common sense to arrange these clips to successfully complete the task.

The final output should be in this format:

Correct order: <mention the Clip numbers separated by a comma>
\end{prompt}
}
\end{minipage}
\begin{minipage}{0.66\linewidth}
\centering
\texttt{\tiny <Video>}
\includegraphics[width=\linewidth]{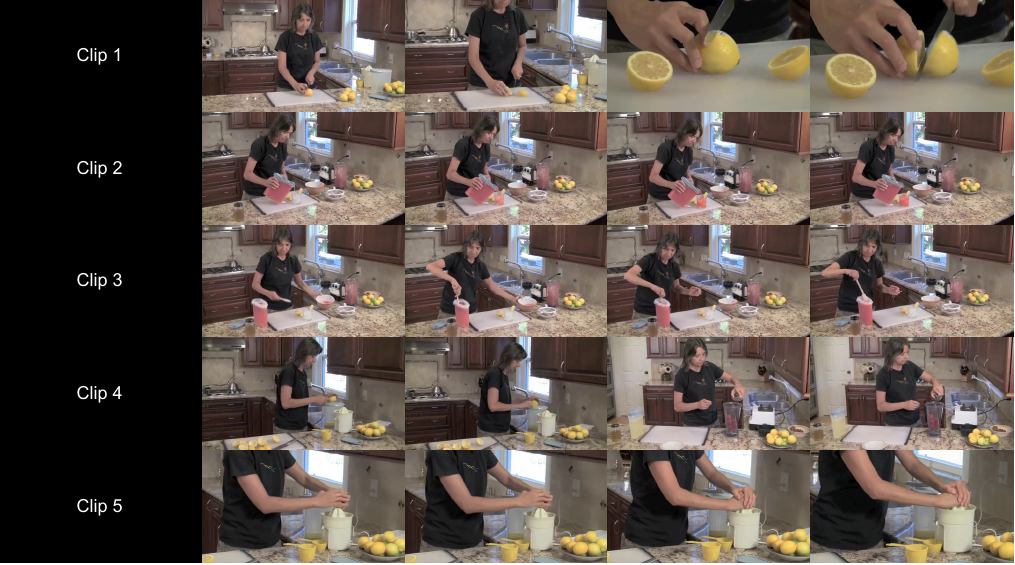}
\end{minipage}
\vspace{0.05in}
\captionof{figure}{
\textbf{Example question and video.}
We present an example of video-based long-form causal reasoning task from \benchmark. 
\textit{The correct order is: Clip 1: Cut lemon into slices, Clip 5: Squeeze lemon into the pitcher, Clip 4: Pour lemon juice and water into the pitcher, Clip 3: Stir the lemonade mixture, Clip 2: Pour lemonade into a glass.}
}
\label{fig:example_benchmark}
\vspace{-0.35in}
\end{table}

To study the video-based causal reasoning capabilities of LVLMs, we introduce \textbf{\textit{Video-based long-form Causal Reasoning Benchmark (\benchmark)}}, an evaluation benchmark consisting of procedural videos depicting everyday human activities, such as making lemonade or grilling steak (see Figure~\ref{fig:example_benchmark} for an example). \benchmark is designed to evaluate whether LVLMs can identify and reason about visual events with long-form causal dependencies towards a specific goal. Specifically, when presented with a shuffled sequence of video clips each showing a key action, the model must first interpret 
the actions in each clip and then arrange them in the correct chronological order based on their causal dependencies to complete the procedure. Unlike prior benchmarks~\cite{intentqa,egoplan,chen2024rextime,li2022representation}, \benchmark explicitly tests multi-step causal reasoning and fine-grained spatio-temporal understanding without allowing linguistic shortcuts common in multiple-choice or binary QA formats.
For instance, in the lemonade-making example (Figure \ref{fig:example_benchmark}), the model must first distinguish between fine-grained actions such as cutting and squeezing a lemon, and subsequently infer that cutting the lemon, squeezing it, and pouring the juice into a pitcher should occur in the correct causal sequence. Our evaluation across both open- and closed-source models shows that current LVLMs struggle with video-based long-form causal reasoning as most perform at or below random guess, and even the best models fall short of human performance by nearly $40\%$. Further analysis reveals that while these models can often recognize individual actions, they frequently fail to establish meaningful connections across a sequence of visual events, lacking an understanding of causal dependencies based on visual observations.

\begin{wraptable}{r}{0.4\textwidth}
\vspace{-0.15in}
\centering
\includegraphics[width=0.8\linewidth]{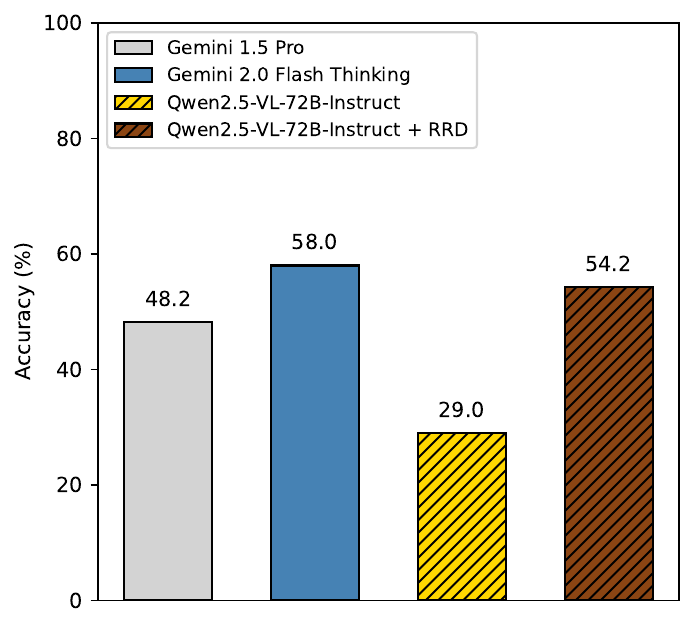}
\vspace{-0.05in}
\captionof{figure}{
\textbf{Impact of RRD.} 
Qwen2.5-VL-Instruct$_\text{72B}$ with RRD outperforms  Gemini-1.5-Pro and achieve comparable performance to Gemini-2-Flash-Thinking.}
\label{fig:cocp_bar_plot}
\vspace{-0.15in}
\end{wraptable}
To improve the long-form causal reasoning capabilities of LVLMs 
we introduce \textbf{\textit{Recognition-Reasoning Decomposition (RRD)}}, a simple modular approach designed to enhance the video-based reasoning abilities of LVLMs. RRD breaks down video-based long-form causal reasoning into two interconnected sub-tasks: (\textit{i}) video recognition and (\textit{ii}) causal reasoning. This decomposition simplifies the overall task by first directing the model’s focus toward recognizing visual events, and then reasoning about their relationships to infer the correct causal order. RRD leads to significant gains, improving accuracies by up to $25.2\%$ on \benchmark. Notably, Qwen2.5-VL$_\text{72B}$-Instruct, when equipped with RRD, surpasses Gemini-1.5-Pro and achieves a performance comparable to the current top-performing reasoning-specialized model, Gemini-2-Flash-Thinking (see Figure~\ref{fig:cocp_bar_plot}).

In summary, our contributions are as follows: 
\begin{itemize}[noitemsep,nolistsep,leftmargin=10pt]
\item 
We introduce \textbf{\benchmark}, a novel benchmark designed to evaluate LVLMs on video-based long-form causal reasoning. 
To the best of our knowledge, this is the first video evaluation benchmark to study multi-step causal reasoning capabilities of LVLMs. 
Our analysis on various state-of-the-art LVLMs reveals that current LVLMs struggle with long-form causal reasoning due to their inability of meaningfully connect a series of visual events toward a goal. 

\item
To improve the performance of open-source LVLMs on \benchmark, we introduce \textbf{RRD}, which decomposes video-based causal reasoning into two related sub-tasks video recognition and causal reasoning. 
This simple modular approach allows LVLMs to focus on one type of task at a time, first recognition, then reasoning, which results in notable performance gains of up to $25.2\%$. 

\end{itemize}

\section{Background}
\textbf{Large Video Language Models (LVLMs).}
LVLMs typically consist of a vision encoder, a Large Language Model (LLM), and a cross-modal adapter that bridges visual and textual modalities~\cite{longvu,slowfast_video,videochat2,llamavid,videollama2,videollama3,timechat,longvila,llavanextvid,videollava,gemini15,internvl,internvl25,qwenvl,qwenvl25}. While this high-level structure is common, recent work has introduced considerable architectural variations. 
These include extending the LLM’s context window for long sequences~\cite{longva,longvila}, dynamic projection techniques that drop redundant frames based on visual similarity~\cite{longvu,moviechat}, and query-based projectors that selectively attend to relevant visual content~\cite{videochat2,internvideo2,videochat_flash,video_salmon}.
In addition to architectural differences, LVLMs vary in their use of vision encoders, ranging from single to multi-encoder setups (e.g., video + image)~\cite{videogptplus}, and from vision-only to vision-language pretrained models~\cite{longvu}. Training strategies also differ, with some models trained in a single stage, and others using multi-stage pipelines that separate large-scale pretraining (for modality alignment) from instruction tuning or reasoning-specialized post-training~\cite{video_llm_survey,video_llm_survey2}.
To ensure a comprehensive evaluation of LVLM capabilities across diverse architectural and pretraining paradigms, we have carefully selected models that represent a broad spectrum within these categories for evaluating on \benchmark.

\textbf{Video evaluation benchmarks.}
Numerous evaluation benchmarks exist for video \textit{understanding} tasks, focusing on areas such as information retrieval-based question answering (e.g., ActivityNetQA \cite{activitynet}, MSRVTTQA \cite{msrvtt_msvd}, MSVDQA \cite{msrvtt_msvd}, NextQA \cite{nextqa}, TGIFQA \cite{tgif}), comprehensive video understanding (e.g., MVBench~\cite{videochat2}, TVBench~\cite{tvbench}, VideoMME \cite{videomme}), fine-grained temporal understanding (e.g., TVBench~\cite{tvbench}, TempCompass \cite{tempcompass}, TemporalBench \cite{temporalbench}), long-video understanding (e.g., MLVU \cite{mlvu}, LongVideoBench \cite{longvideobench}), egocentric video understanding (e.g., Egoschema \cite{egoschema}), and video hallucination (e.g., VideoHallucer \cite{videohallucer}, HallusionBench \cite{hallusionbench}), among others. There also exist a few benchmarks focused on video-based reasoning, such as SOK-Bench \cite{wang2024sok}, MMWorld \cite{he2024mmworld}, and VILMA \cite{kesen2023vilma}. However, a significant gap remains in the evaluation of video-based \textit{causal reasoning} tasks.
While some benchmarks address intent (e.g., IntentQA \cite{intentqa}), causal question answering (e.g., Causal-VidQA \cite{li2022representation}), or goal-oriented question answering (e.g., EgoPlan-Bench \cite{egoplan}, ReXTime \cite{chen2024rextime}), they do not adequately assess the video-based \textit{long-form} or \textit{multi-step} causal reasoning capabilities of LVLMs. 
In this work, we address the critical area of long-form causal reasoning, which refers to reasoning about visual events with multiple or interconnected causal dependencies.

\textbf{Reasoning methods.} 
Chain-of-Thought prompting has emerged as a powerful technique to improve reasoning in LLMs and LVLMs by encouraging intermediate step-by-step derivations rather than direct answer prediction \cite{cot}. This paradigm has been further strengthened by post-training alignment techniques such as Reinforcement Learning with Human Feedback (RLHF) \cite{rlhf,ppo}, which optimize models to generate more helpful and aligned responses. More recent methods like DeepSeek's R1 \cite{deepseek_r1} also build on such alignment strategies to enhance reasoning quality. In parallel, a growing body of work explores inference-time techniques to boost performance without the necessity of additional training. These include majority voting or self-consistency sampling \cite{wang2022self}, which aggregate multiple generated responses for robustness, best-of-N sampling \cite{stiennon2020learning,nakano2021webgpt}, which selects the highest-quality sample from multiple candidates, and decomposed prompting \cite{khot2022decomposed}, which breaks complex reasoning tasks into simpler sub-tasks. Our proposed approach, RRD, is motivated by decomposed prompting where complex video-based reasoning tasks are systematically divided into several sub-tasks.

\section{Video-based long-form Causal Reasoning Benchmark (\benchmark)} 

\subsection{Construction of \benchmark}
We construct \benchmark by curating a set of everyday procedures that require no specialized knowledge and are commonly encountered in daily life, such as grilling steak, making lemonade, or preparing pancakes (see Appendix~\ref{suppsec:benchmark_details} for the full list). For each procedure, we source instructional videos from the CrossTask dataset~\cite{crosstask}, which contains YouTube videos with human annotated timestamps of key events. Below, we outline the three-stage process for preparing videos and questions in \benchmark.

\noindent \textbf{Preparing the videos.}
Our video construction pipeline consists of the following steps:

$\bullet$ \textbf{Step 1.} We begin with a complete procedural video and use the provided human-annotated timestamps to segment it into short clips, each corresponding to a specific procedural step (e.g., \textit{seasoning steak} for the procedure \textit{grill steak}).   

$\bullet$ \textbf{Step 2.} Using WikiHow as a reference (\url{https://www.wikihow.com/}), we identify the core steps necessary for the procedure. We group consecutive steps that have no causal dependencies. Moreover, we remove irrelevant segments that do not contribute to the main task, ensuring that all selected clips exhibit causal dependencies. At this stage, the resulting set of clips must follow a meaningful chronological order for successful completion of the procedure. This step is manually curated by human annotators to ensure accurate assessment.

$\bullet$ \textbf{Step 3.} The selected clips are then randomly shuffled, with the constraint that the original ascending order is not retained. The shuffled clips are concatenated into a single video, with blank frames inserted between them for visual separation. The blank frames preceding the clips labeled chronologically to clearly distinguish the individual steps. The resulting video serves as the input to the LVLM, which is tasked with identifying the correct order of the procedural steps. 

\begin{figure}[t]
    \centering
    \includegraphics[width=\linewidth]{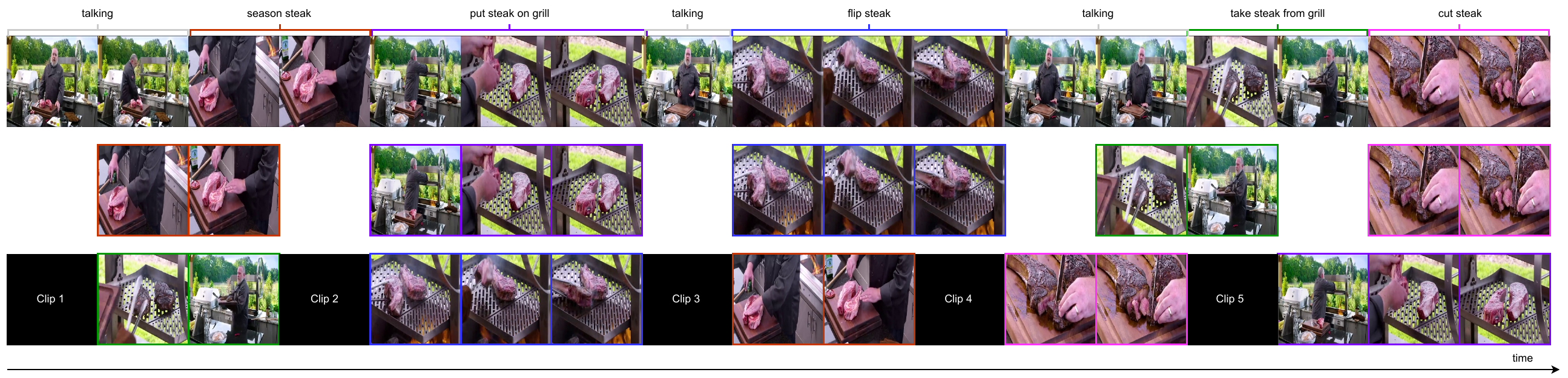}
    \caption{    
    \textbf{Overview of video construction.}
    \textbf{Step 1:} Given a complete video, key procedural steps are identified based on human-annotated timestamps.
    \textbf{Step 2:} We keep the key events and discard those that do not depict visual events directly associated with the goal, such as talking or narrating in this example of grilling steak.
    \textbf{Step 3:} Each key event is shuffled across time and assigned a clip number. These clips are then merged together to form the final test sample.    
    }
    \label{fig:full_process}

\vspace{0.1in}
\resizebox{\linewidth}{!}{
\begin{prompt2}
The given video consists of multiple short clips, each showing a different segment needed to complete the task: \blue{\{name of the procedure\}}. 

These clips are randomly shuffled, and your job is to arrange them in the correct order to complete the task: \blue{\{name of the procedure\}}. 

The clip numbers are mentioned at the beginning of each clip as Clip 1, Clip 2, and so on.

In order to solve this task, first, you should identify the activity that is performed in each clip, and then use your reasoning and common sense to arrange these clips to successfully complete the task.

The final output should be in this format:

Correct order: <mention the Clip numbers separated by a comma>
\end{prompt2}
}
\captionof{figure}{The \textbf{question template} used in \benchmark.}
\label{fig:pp_prompt_video_only}
\vspace{-0.25in}
\end{figure}

An overview of this construction process is illustrated in Figure~\ref{fig:full_process}.

\begin{figure}%
\centering
\renewcommand{\arraystretch}{1}  
\fontsize{9pt}{10pt}\selectfont
\resizebox{\linewidth}{!}{
\begin{tabular}{ccc}
\includegraphics[height=0.38\linewidth]{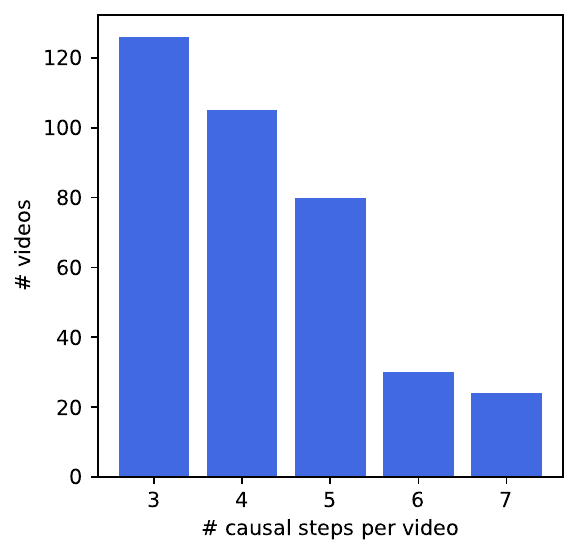}
&  
\includegraphics[height=0.38\linewidth]{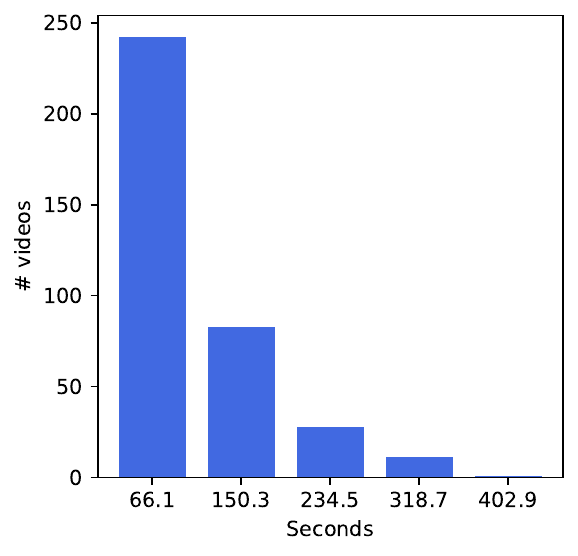}
&
\includegraphics[height=0.38\linewidth]{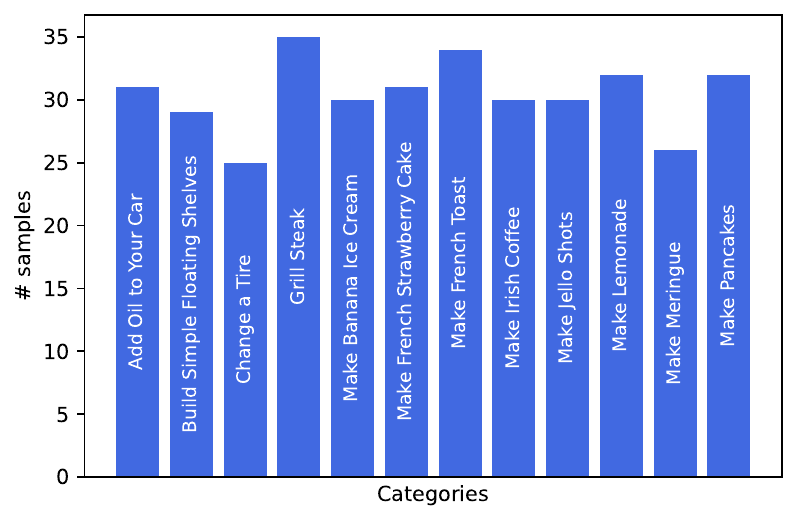}\\
\small{(A) Distribution of causal steps.} & \small{(B) Distribution of video durations.} & \small{(C) Distribution of procedures.}\\
\multicolumn{3}{c}{
\begin{tabular}{cc}
\includegraphics[height=0.36\linewidth]{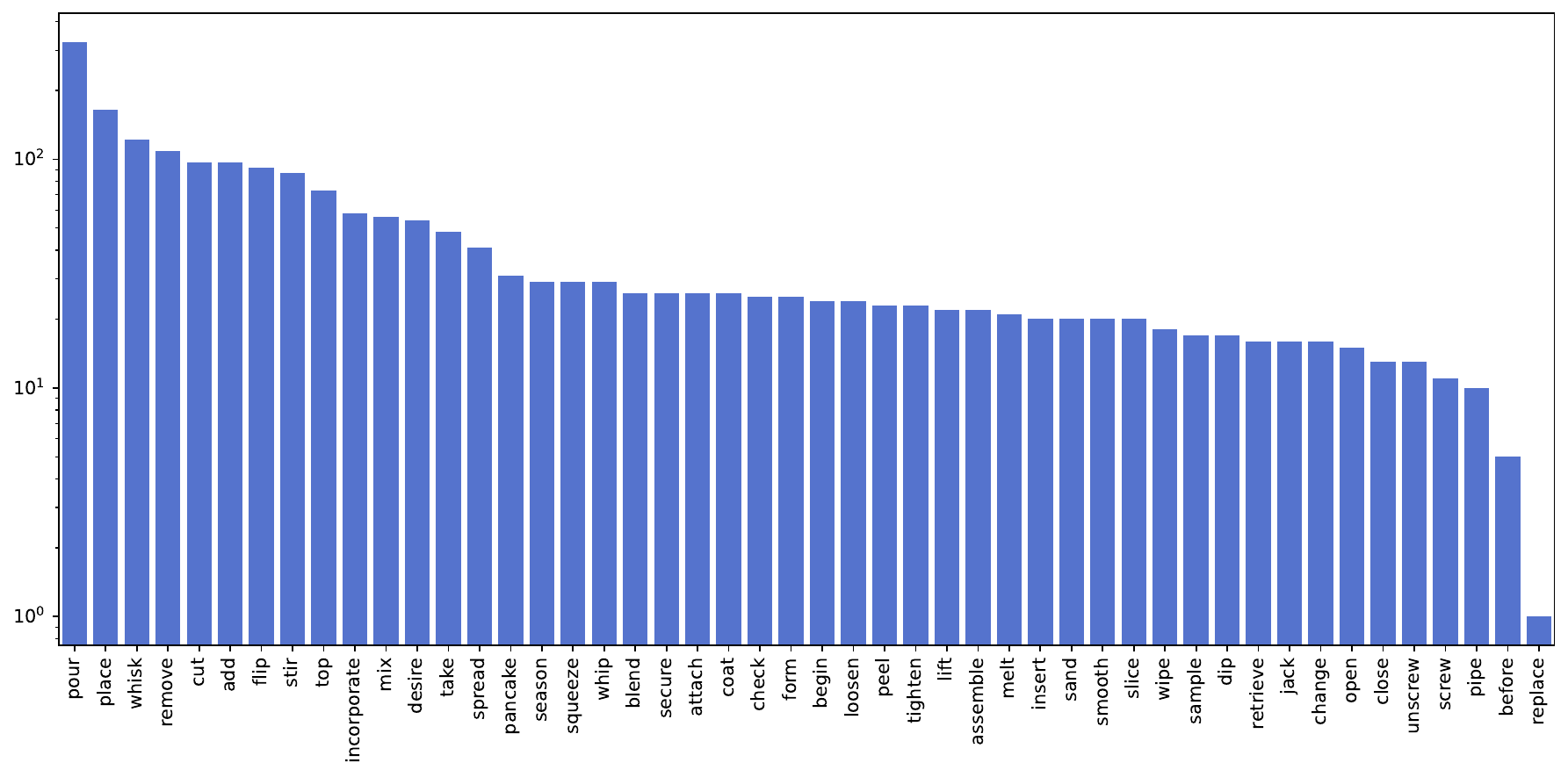} & 
\includegraphics[height=0.36\linewidth]{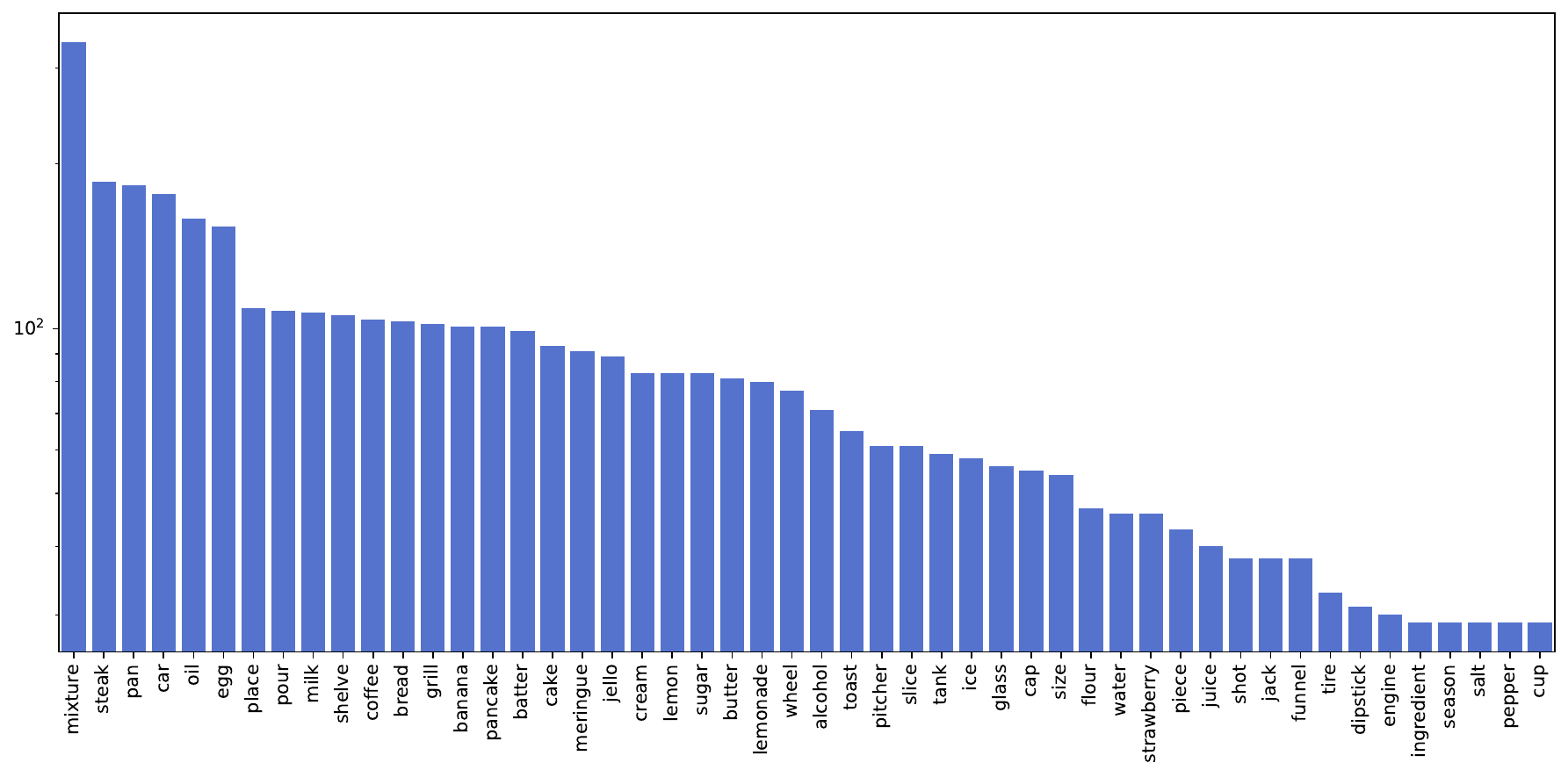}\\
\small{(D) Top-50 fine-grained actions.} & \small{(E) Top-50 objects.}\\
\end{tabular}
}
\end{tabular}
}
\captionof{figure}{
Key \textbf{statistics} of our \benchmark.
}
\label{fig:benchmark_stat}
\vspace{-0.15in}
\end{figure}

\textbf{Preparing the questions.}
A key challenge in evaluating LVLMs is designing a reliable evaluation protocol to correctly asses true visual understanding. Most existing video benchmarks~\cite{videochat2,tvbench,videohallucer,videomme,longvideobench,tempcompass,activitynet,msrvtt_msvd,nextqa,tgif,egoschema,hallusionbench,egoplan,intentqa} rely on multiple-choice or binary question-answering formats, to streamline their automated evaluation. However, these formats can be exploited through linguistic cues in the provided response choices, without requiring true visual understanding. Open-ended question answering offers a more rigorous probe of visual reasoning, but introduces evaluation ambiguity which often necessitates the use of an external LLM (e.g., GPT-4~\cite{gpt4}) as a judge for automated evaluation, a strategy proven to be unreliable and ambiguous in prior work~\cite{tvbench}.
\benchmark addresses these limitations by framing causal reasoning as a sequence ordering task. This setup avoids the use of linguistic cues in predefined options, yet yields deterministic ground truth answers. As a result, it enables accurate, objective evaluation while still challenging LVLMs to perform fine-grained visual and causal reasoning. The question template is mentioned in Figure \ref{fig:pp_prompt_video_only}.

\textbf{Statistics.}
\benchmark comprises $365$ videos and questions
across $12$ categories of procedures covering diverse fine-grained actions and object interactions.
The videos are $30$ to $445$ seconds long with an average duration of $107$ seconds and a total duration of $10$ hours. To keep the difficulty reasonable, we include videos requiring only $3$ to $7$ causally dependent steps with an average of $4.2$ steps per task. Additional key statistics are provided in Figure \ref{fig:benchmark_stat}.

\subsection{Evaluation Metrics}
We measure the performance of LVLMs on \benchmark with two metrics: overall accuracy and step accuracy \cite{chang2020procedure}. Overall accuracy (also referred to simply as Accuracy) indicates predictions that exactly match the ground truth, whereas step accuracy compares predicted and ground truth actions step by step. Assume, $(q, v) \in \gD$, where $q$ is a question related to a video $v$ sampled from a validation set $\gD$. Let $\pi$ be an LVLM and $\mathbbm{1}(\cdot)$ be the indicator function of correct prediction.
The mathematical expressions of our evaluation metrics are as follows:
\begin{equation*}
    \text{Accuracy}\!=\!\frac{\sum_{(q, v) \in \gD} \mathbbm{1}\left(\pi(q, v)\right)}{|\gD|}
    ,
    \quad \quad
    \text{Step Accuracy}\! =\! \frac{\sum_{(q, v) \in \mathcal{D}} \frac{\sum_{s} \mathbbm{1}\left(\pi(q, v)\right)}{s}}{|\mathcal{D}|},
\end{equation*}
where $s$ denotes the total number of steps to a procedure.

\section{Benchmarking Results}
\vspace{-0.05in}

\subsection{Setup}
We examine over $20$ recent and popular LVLMs, including both closed and open-source models. %
These models exhibit significant variations in several key aspects: LLM architectures (LLaMA \cite{llama1,llama2,llama3}, Mistral \cite{mistral}, and Qwen \cite{qwen,qwen2}) with sizes from 1B to 78B parameters for open-source LVLMs; cross-modal adapters (QFormer \cite{blip2}, MLP projector \cite{llava_improved}, and spatio-temporal compressor \cite{longvu,videollama2}); vision encoders with single or dual configurations (CLIP \cite{clip}, SigLIP \cite{siglip}, DINO \cite{dinov2}, and UMT \cite{umt}); training methodologies (single-stage or multi-stage) including alignment finetuning for improved reasoning (\cite{qwenvl25,rrpo}); and visual frame processing capabilities ranging from $8$ (NVILA \cite{nvila}) to over $500$ frames (LongVU \cite{longvu}, Qwen2.5-VL \cite{qwenvl25}). We follow the recommended generation configurations, such as temperature, system prompt, number of frames, and other key parameters, for each respective LVLM. For reference, we also benchmark human performance on \benchmark.

\subsection{Results and Findings}
Here, we discuss our key observations regarding the performance of LVLMs on \benchmark, based on our detailed quantitative and qualitative analysis.

\textit{\textbf{\darkgreen{VCRBench tasks are unambiguous to human evaluators.}}}
As shown in Table~\ref{tab:benchmarking_pp}, human participants achieve an accuracy of $96.4\%$ on \benchmark. This high performance indicates that the video-based long-form causal reasoning tasks are intuitive and unambiguous to humans. Further details on the human evaluation setup are provided in Appendix~\ref{suppsec:benchmark_details}.

\begin{table*}[]
\centering
\begin{minipage}{0.48\textwidth} %
\centering
\fontsize{9pt}{10pt}\selectfont
\setlength\tabcolsep{1pt}
\caption{\textbf{Results on \benchmark.} 
Most open-source LVLMs perform at or below random guess, and even the best LVLM falls significantly short of human performance. 
We \fadetext{fade} numbers that fall below the random guess baseline. 
}
\label{tab:benchmarking_pp}
\resizebox{\linewidth}{!}{
    \begin{tabular}{lccc}
    \toprule
    \bf Models & \bf \# Frames & \bf Acc.$_{(\uparrow)}$ & \bf Step Acc.$_{(\uparrow)}$
    \\ \midrule\midrule
    Random Guess & & 7.8 & 24.1 \\
    \midrule
    InternVL2.5$_\text{1B}$ \cite{internvl25} & $64$ & $\fadetext{1.4}$ & $\fadetext{10.3}$\\
    InternVL2.5$_\text{2B}$ \cite{internvl25} & $64$ & $\fadetext{6.3}$ & $\fadetext{16.2}$\\
    LongVU$_\text{3B}$ \cite{longvu} & $1$fps & $\fadetext{0.0}$ & $\fadetext{7.0}$\\
    InternVL2.5$_\text{4B}$ \cite{internvl25} & $64$ & $\fadetext{1.6}$ & $\fadetext{9.5}$\\
    VideoChat2$_\text{7B}$ \cite{videochat2} & $16$ & $\fadetext{0.3}$ & $\fadetext{5.8}$\\
    InternVL2.5$_\text{8B}$ \cite{internvl25} & $64$ & $\fadetext{2.7}$ & $\fadetext{11.1}$\\
    LLaVA-NeXT-Video$_\text{7B}$ \cite{llavanextvid} & $64$ & $\fadetext{0.0}$ & $\fadetext{17.4}$\\
    MiniCPM-o-V 2.6$_\text{7B}$ \cite{minicpm_v} & $64$ & $\fadetext{2.5}$ & $\fadetext{11.0}$\\
    Qwen2.5-VL-Instruct$_\text{7B}$ \cite{qwenvl25} & $1$fps & $\fadetext{7.1}$ & $\fadetext{20.9}$\\
    VideoLLaMA3$_\text{7B}$ \cite{videollama3} & $128$ & $\fadetext{1.6}$ & $\fadetext{13.1}$\\
    LongVILA$_\text{7B}$  \cite{longvila} & $128$ & $\fadetext{0.3}$ & $\fadetext{1.1}$\\
    LongVU$_\text{7B}$ \cite{longvu} & $1$fps & $\fadetext{0.0}$ & $\fadetext{2.4}$\\
    NVILA$_\text{15B}$ \cite{nvila} & $8$ & $\fadetext{0.6}$ & $\fadetext{3.6}$\\
    InternVL2.5$_\text{26B}$ \cite{internvl25} & $64$ & $\fadetext{2.7}$ & $\fadetext{13.7}$\\
    InternVL2.5$_\text{38B}$ \cite{internvl25} & $64$ & $11.0$ & $27.4$\\
    LLaVA-NeXT-Video$_\text{72B}$ \cite{llavanextvid} & $32$ & $\fadetext{5.2}$ & $\fadetext{18.6}$\\
    \bronzeicon Qwen2.5-VL-Instruct$_\text{72B}$ \cite{qwenvl25} & $1$fps & $29.0$ & $44.0$\\
    InternVL2.5$_\text{78B}$ \cite{internvl25} & $64$ & $14.5$ & $34.0$\\ %
    GPT4o \cite{gpt4o} & $32$ & $29.0$ & $36.6$\\
    \silvericon Gemini-1.5-Pro \cite{gemini15} & $1$fps & $48.2$ & $65.3$\\
    \goldicon Gemini-2.0-Flash-Thinking \cite{gemini15} & $1$fps & $58.0$ & $67.7$\\
    \midrule
    Human & & $\mathbf{96.4}$ & $\mathbf{98.3}$ \\
    \bottomrule
    \end{tabular}
}
\end{minipage}%
\hfill
\begin{minipage}{0.48\textwidth} %
\centering
\begin{prompt}
    \blue{\# Response from LongVILA} 
    
    Correct Order: Clip 1, Clip 2, ..., Clip 14, ...
    
    -----------------------------------------------------------------------------------------------------------------
    
    \blue{\# Response from LongVU} 
    
    Correct Order: 1, 2, 3, 4, 5, ..., 29, 30, ...
    
    -----------------------------------------------------------------------------------------------------------------
    
    \blue{\# Response from LLaVA-Next-Qwen} 
    
    Correct Order: Clip 1, Clip 2, Clip 3, Clip 4
    
    \end{prompt}
\vspace{-0.15in}
\captionof{figure}{
\textbf{Failure examples.}  
Several open-source LVLMs merely list consecutive numbers as the predicted order, exhibiting inability to make a meaningful attempt in \benchmark tasks.}
\label{fig:open_source_degeneration}

\includegraphics[width=\linewidth]{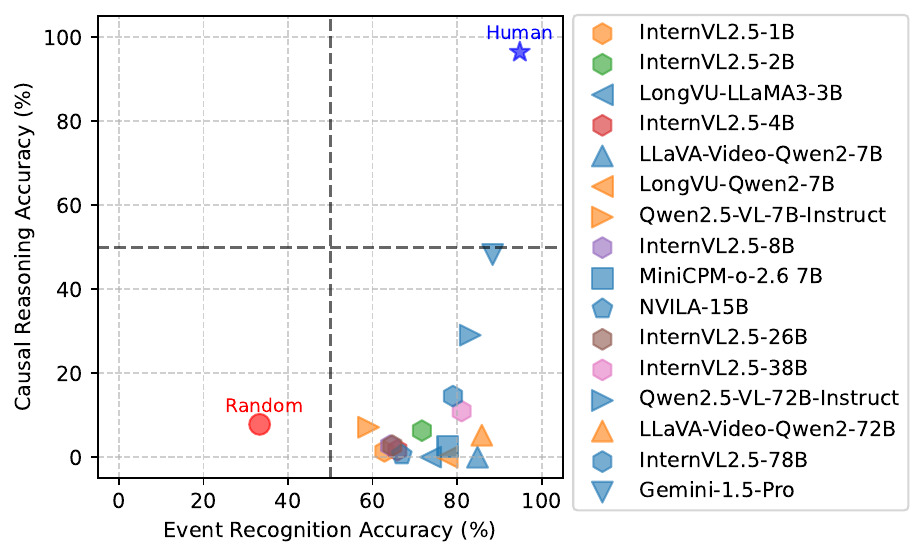}
\vspace{-0.15in}
\captionof{figure}{
\textbf{Reasoning vs. Recognition.} 
LVLMs can identify individual events but fail to connect them toward a specific goal in causal reasoning.}
\label{fig:pp_anal}
\end{minipage}
\vspace{-0.15in}
\end{table*}

\input{figures/failed_example}

\textit{\textbf{\darkgreen{Recognizing events is not enough: LVLMs lack associative understanding of visual events.}}} To better understand the limitations of current LVLMs, we conduct additional experiments that evaluate their ability to recognize intermediate events, which is a prerequisite for long-form causal reasoning. We design a multiple-choice question-answering (MCQ) evaluation where each sample presents a clip of an intermediate step along with three answer options: one correct and two incorrect. We derive the incorrect options from the other steps within the procedure. Figure~\ref{fig:pp_anal} plots video recognition accuracy against causal reasoning accuracy to analyze their relationship. As shown in Figure~\ref{fig:pp_anal}, most LVLMs achieve decent accuracy ($\sim$$80\%$) on event recognition, while performing poorly in long-form causal reasoning tasks. This suggests that while LVLMs can recognize individual visual events when provided answer choices, they struggle with video recognition in an open-ended setup and fail to connect these events meaningfully toward a goal. Our results further suggest that recognizing events is necessary but not sufficient for long-form causal reasoning. For example, although LLaVA-Video-Qwen2$_\text{72B}$ performs better than Qwen2.5-VL-Instruct$_\text{72B}$ on the video recognition task, it completely fails at causal reasoning. 

\section{A Simple Step Towards Improving Video-based Causal Reasoning}

\subsection{Recognition-Reasoning Decomposition} 
Humans excel at reasoning by decomposing complex tasks into a series of sub-tasks, addressing each in a sequential manner, and leveraging the intermediate results to arrive at a final conclusion. Inspired by this cognitive problem-solving strategy, we propose a modular approach that explicitly decomposes video-based causal reasoning tasks into two distinct, yet interdependent, sub-problems: (\textit{i}) video recognition, which aims to extract salient events from the visual input, and (\textit{ii}) causal reasoning, which involves inferring the causal relationships between these identified events. This decomposition allows the LVLM to focus on one type of task at a time. We refer to this approach as Recognition-Reasoning Decomposition (RRD).
The details of the sub-tasks are described as follows:

$\bullet$ \textbf{Video recognition.} 
As the first stage of our approach, we instruct the LVLM to obtain the descriptions of fine-grained actions/events for each clip of the video using the following prompt: \\
\texttt{Provide a one-sentence description indicating the key and fine-grained actions or events for each clip. Please respond in this format:\\
Clip 1: <Write one sentence description>\\
Clip 2: <Write one sentence description>.}\\
This allows the LVLM to strictly focus on the actions and events without necessarily considering the causal relationships among clips, enabling explicit focus on and localized analysis of each clip.

$\bullet$ \textbf{Causal reasoning.} 
The next stage of RRD involves arranging the identified events from the video recognition step based on their causal relationships to complete the procedure. Note that the clips are shuffled, and thus, so are the identified events. 
To this end, we instruct the LVLM to identify the correct order of the events identified in the previous stage, using the following prompt:\\
\texttt{The following randomly shuffled steps are needed to complete the task: \{name of the procedure\}.\\
Use your reasoning and common sense to arrange these steps to successfully complete the task.\\
\{clip descriptions\}
}\\
This process enables the LVLM to leverage its language capabilities for reasoning tasks. %

\subsection{Experiments and Results}

To test RRD on our proposed \benchmark, we use the top-performing open-source LVLMs (based on their performance on \benchmark in Table \ref{tab:benchmarking_pp}), i.e., InternVL2.5 and Qwen2.5-VL-Instruct. Specifically, we use both the $7$B and $72$B variants of Qwen2.5-VL-Instruct and $38$B and $78$B variants of InternVL2.5. 
We follow the recommended inference setup of these LVLMs and use $64$ frames for  InternVL2.5 and sample frames at $1$ FPS for Qwen2.5-VL-Instruct, similar to Table \ref{tab:benchmarking_pp}. 
Following we provide our findings regarding the behavior of RRD on \benchmark, along with detailed experimental results and analysis.

\begin{wraptable}{r}{0.5\linewidth}
\vspace{-0.15in}
\centering
\fontsize{9pt}{10pt}\selectfont
\setlength\tabcolsep{2pt}
\caption{
\textbf{Impact of RRD.}
Our proposed task decomposition significantly enhances the long-form causal reasoning capabilities of LVLMs, yielding accuracy improvements of between $12.6\%$ and $20.9\%$.
}
\label{tab:main_result_rrd}
\begin{tabular}{lcc}
\toprule
\bf Models & \bf Acc.$_{(\uparrow)}$ & \bf Step Acc.$_{(\uparrow)}$ \\
\midrule\midrule
Qwen2.5-VL-Instruct$_\text{7B}$ \cite{qwenvl25} & $7.1$ & $20.9$\\
 ~~~+ \bf RRD (Ours)  & $22.5$\inc{15.4} & $37.3$\inc{16.4}\\
\midrule
InternVL2.5$_\text{38B}$ \cite{internvl25} & $11.0$ & $27.4$\\
 ~~~+ \bf RRD (Ours)  & $23.6$\inc{12.6} & $34.3$\inc{6.9}\\
\midrule
Qwen2.5-VL-Instruct$_\text{72B}$ \cite{qwenvl25} & $29.0$ & $43.0$\\
 ~~~+ \bf RRD (Ours)  & $\mathbf{49.9}$\inc{20.9} & $\mathbf{63.4}$\inc{20.4}\\
\midrule
InternVL2.5$_\text{78B}$ \cite{internvl25} & $14.5$ & $34.0$\\
 ~~~+ \bf RRD (Ours)  & $28.2$\inc{13.7} & $43.5$\inc{9.5}\\
\bottomrule
\end{tabular}
\end{wraptable}%
\textit{\textbf{\darkgreen{RRD significantly improves video-based long-form causal reasoning capabilities of LVLMs.}}}%
The results in Table~\ref{tab:main_result_rrd} demonstrate that our proposed RRD significantly enhances the video-based causal reasoning capabilities of LVLMs. The benefits of RRD are consistent across different model sizes (from 7B to 78B) and across both weaker to stronger LVLMs. 
For instance, Qwen2.5-VL-Instruct$\text{7B}$, which initially performed close to random guess on \benchmark, achieves a $15.3\%$ accuracy gain when equipped with RRD. Similarly, the 38B and 78B variants of InternVL2.5 show improvements of $12.6\%$ and $13.7\%$, respectively. Moreover, the top-performing open-source LVLM Qwen2.5-VL-Instruct$\text{72B}$ sees a substantial improvement of $20.8\%$ in accuracy when equipped with RRD. 
Note that RRD improves the performance of LVLMs that rely on a fixed number of visual inputs, such as InternVL2.5, as well as models that accept a varying number of frames, such as Qwen2.5-VL-Instruct.

\begin{wraptable}{r}{0.5\textwidth}
\vspace{-0.15in}
    \centering
    \fontsize{9pt}{10pt}\selectfont
    \setlength\tabcolsep{2pt}
    \caption{
    The effect of incorporating videos at causal reasoning step. The results are based on Qwen2.5-VL-Instruct$_\text{72B}$. 
    Here \faVideoCamera~ refers to videos and
    \faFileText~ refers to generated video descriptions from video recognition step. Using videos at reasoning stage degrade performance.
    }
    \label{tab:abl_result_videos}
    \vspace{-0.05in}
    \begin{tabular}{cccccc}
    \toprule
    \specialcell{\bf Video\\\bf Recognition} &
    \specialcell{\bf Causal\\\bf Reasoning} &
    \bf Acc.$_{(\uparrow)}$ & \bf Step Acc.$_{(\uparrow)}$ \\ \midrule\midrule
    \faVideoCamera & \faFileText & $49.9$ & $63.4$\\
    \faVideoCamera & \faVideoCamera~+ \faFileText & $46.6$\decbad{3.3} & $62.8$\decbad{0.6}\\
    \bottomrule
    \end{tabular}
\vspace{-0.15in}
\end{wraptable}
\textit{\textbf{\darkgreen{LVLMs mainly depend on their language knowledge for complex reasoning while including vision may hinder performance.}
}}
The results presented in Table \ref{tab:abl_result_videos} suggest that LVLMs mainly rely on their language abilities when solving complex reasoning tasks. Interestingly, we find that incorporating videos in addition to the clip descriptions at the causal reasoning step degrades the accuracy of LVLMs on \benchmark. 
This performance drop may be due to possible conflicts or misalignment between the visual and linguistic understanding of the model.

\begin{wraptable}{r}{0.5\textwidth}
    \centering
    \fontsize{9pt}{10pt}\selectfont
    \caption{
    The effect of further decomposing video recognition and causal reasoning steps. 
    }
    \label{tab:abl_result_iter}
    \begin{tabular}{llcc}
    \toprule
    \specialcell{\bf Video\\\bf Recognition} &
    \specialcell{\bf Causal\\\bf Reasoning} &
    \bf Acc.$_{(\uparrow)}$ & \bf Step Acc.$_{(\uparrow)}$ \\
    \midrule\midrule
    all-at-once & all-at-once  & $49.9$ & $63.4$\\
    all-at-once & sequential  & $47.4$ & $64.1$\\
    sequential & all-at-once  & $\mathbf{54.2}$ & $65.1$\\
    sequential & sequential  & $51.0$ & $\mathbf{66.6}$\\
    
    \bottomrule
    \end{tabular}
\vspace{-0.15in}
\end{wraptable}
\textit{\textbf{\darkgreen{Sequential recognition improves performance by focusing on one key event at a time, while sequential reasoning degrades overall accuracy due to the need for global context in long-form causal reasoning.}}}
Next, we conduct a thorough analysis in the main design of RRD by
examining the effect of performing video recognition across all clips (referred to as \textit{all-at-once}) versus analyzing each clip individually (referred to as \textit{sequential}). %
Intuitively, the sequential approach further simplifies the video recognition task and allows the LVLM to focus on localized analysis of one clip at a time. 
The detailed setup for this experiment is presented in Appendix \ref{suppsec:rrd}. 
Similarly, for causal reasoning, we explore  pairwise causal comparisons in a sequential manner against determining the correct causal order all at once. 
To perform pairwise comparisons, we adopt a sorting algorithm (i.e., \text{merge sort}) that arranges visual events into a causal chain, where each comparison is based on the causality between two events as determined by the LVLM. Upon completion of sorting, the resulting ordered list of events is used as the final prediction. The detailed setup for this experiment is provided in Appendix \ref{suppsec:rrd}. We conduct this experiment utilizing the best open-source LVLM Qwen2.5-VL-Instruct$_\text{72B}$. 
The results are presented in Table~\ref{tab:abl_result_iter}, which reveal the following: (\textit{i}) performing video recognition sequentially helps LVLMs focus on one key event at a time, leading to improved accuracy in \benchmark; (\textit{ii}) for causal reasoning, however, the sequential approach does not yield better results. We conjecture that this is due to the long-range dependencies among causal events: presenting all events together allows the LVLM to better capture the global causal structure, whereas pairwise comparisons provide only local views. Although step accuracy, which measures the correctness of individual steps, shows slight improvement, the overall reasoning accuracy is lower in the sequential causal reasoning setup compared to the all-at-once approach.

\begin{table}[!h]
\vspace{-0.15in}
\centering
\fontsize{9pt}{10pt}\selectfont
\setlength\tabcolsep{2pt}
\caption{Qwen2.5-VL-Instruct$_\text{72B}$ equipped with RRD outperforms Gemini-1.5-Pro and achieve a comparable performance to Gemini-2.0-Flash Thinking.  
}
\label{tab:detailed_results_rrd}
\resizebox{0.8\linewidth}{!}{
\begin{tabular}{lcccccc}
\toprule
\multirow{2}{*}{\bf Models}
& \multicolumn{5}{c}{\bf Number of causal steps}
& \multicolumn{1}{c}{\bf Overall} \\
\cmidrule(lr){2-6} \cmidrule(lr){7-7}
& \bf 3 
& \bf 4
& \bf 5 
& \bf 6
& \bf 7
& \bf Acc.  \\
\midrule\midrule

Qwen2.5-VL-Instruct$_\text{72B}$ \cite{qwenvl25} & $40.5$ & $37.1$ & $16.2$ & $6.7$ & $4.2$ & $29.0$\\
 ~~~+ \bf RRD (ours)  & $64.3$\inc{19.8} & $\mathbf{73.3}$\inc{36.2} & $31.2$\inc{15.0} & $23.3$\inc{16.6} & $\mathbf{33.3}$\inc{29.1} & $54.2$\inc{25.2} \\
\midrule
Gemini-1.5-Pro \cite{gemini15} & $60.3$ & $50.5$ & $36.2$ & $43.3$ & $20.8$ & $48.2$\\
Gemini-2.0-Flash-Thinking \cite{gemini15} & $\mathbf{64.8}$ & $66.7$ & $\mathbf{46.2}$ & $\mathbf{53.3}$ & $29.2$ & $\mathbf{58.0}$ \\

\bottomrule
\end{tabular}
}
\vspace{-0.15in}
\end{table}
\textbf{\textit{\darkgreen{RRD effectively enhances video-based causal reasoning across varying number of steps.}
}}
Table~\ref{tab:detailed_results_rrd} shows that the benefits of RRD are consistent across videos with varying number of steps. Notably, Qwen2.5-VL-Instruct$_\text{72B}$ equipped with RRD outperforms Gemini-1.5-Pro by $6\%$ and achieves a performance comparable to the reasoning-focused Gemini-2.0-Flash-Thinking. In some setups, it even surpasses Gemini-2.0-Flash-Thinking (see Table~\ref{tab:detailed_results_rrd} for details).

\section{Discussions}

\textbf{Summary.} 
In this work, we introduce \benchmark, a novel benchmark designed to evaluate video-based long-form causal reasoning capabilities of LVLMs. Through a comprehensive study of over $20$ recent and popular LVLMs, we find that current models consistently struggle with long-form causal reasoning based on visual observations, largely due to a lack of associative understanding of visual events. As an initial step towards enabling such capabilities we introduce RRD, a simple approach that decomposes video-based causal reasoning into video recognition and causal reasoning tasks. RRD significantly improves long-form causal reasoning capabilities of LVLMs, for instance, Qwen2.5-VL-Instruct$\text{72B}$ with RRD outperforms Gemini-1.5-Pro and achieves performance comparable to Gemini-2-Flash-Thinking. 

\textbf{Limitations and future work.}
While RRD effectively improves the video-based causal reasoning capabilities of LVLMs, it relies on explicit human instructions for decomposing complex reasoning tasks. Future work could focus on developing LVLMs that can internalize such modular approach in decomposing complex tasks without explicit human guidance. 
Furthermore, future work could focus on developing LVLMs capable of performing complex reasoning directly from visual inputs, in contrast to current models that primarily rely on language knowledge for such tasks.

\clearpage

%% file: figures/failed_example.tex
\textit{\textbf{\darkgreen{LVLMs lack video-based long-form causal reasoning abilities.}}}
As shown in Table~\ref{tab:benchmarking_pp}, most open-source LVLMs perform worse than random guessing, with the exception of InternVL$_\text{38B}$, Qwen2.5-VL-Instruct$_\text{72B}$, and InternVL$_\text{78B}$. 
Several open-source LVLMs (e.g., LongVILA, LLaVA-NeXT-Video, LongVU) exhibit a tendency to output a sequence of consecutive numbers, up to their maximum generation length, as the presumed correct order, see examples in Figure~\ref{fig:open_source_degeneration}. This suggests that these models have not developed a robust notion of attempting video-based causal reasoning tasks in \benchmark, and instead default to token-level statistical regularities when uncertain. 
Surprisingly, even open-source models built for improved reasoning, such as MiniCPM-o \cite{minicpm_v}, underperform on \benchmark, suggesting limited video-based causal reasoning abilities. Among open-source models, Qwen2.5-VL-Instruct$_\text{72B}$ performs best, though it still lags significantly behind the best closed-source model, Gemini-2-Flash-Thinking. Interestingly, GPT-4o performs worst among the closed-source models, likely due to its limited capacity for long visual inputs, achieving a performance similar to Qwen2.5-VL-Instruct$_\text{72B}$.
Overall, even the best-performing model, Gemini-2-Flash-Thinking, falls substantially short of human-level performance ($58.0\%$ vs. $96.4\%$). We present sample responses from top-performing LVLMs in Figure \ref{fig:qual_example} and additional results in Appendix \ref{suppsec:benchmark_results}. 

\begin{figure}
\centering
\includegraphics[width=0.975\linewidth]{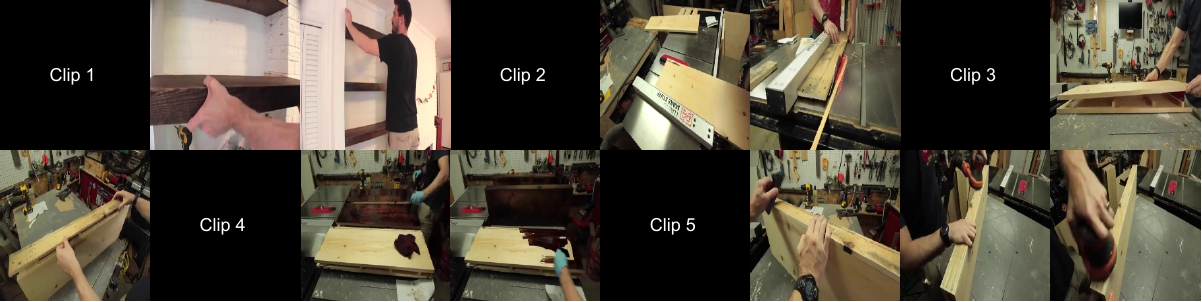}
\resizebox{\linewidth}{!}{
\begin{prompt}

\blue{\# Response from Gemini-2.0-Flash-Thinking}

Rationale:

To build simple floating shelves, the process generally follows these steps:

1. Prepare the wood: This involves cutting the wood to the required sizes and sanding it to create a smooth surface for finishing. Clip 5 shows the sanding process of the wooden components. Although cutting wood to size isn't shown as a dedicated clip, it is implied and precedes assembly. Clip 3 partially shows cutting wood on a table saw and chop saw.

2. Assemble the shelf structure: Once the wood pieces are prepared, the next step is to assemble the shelf frame. Clip 3 demonstrates this step by showing the use of wood glue and a nail gun to join wooden pieces to form the shelf's inner support structure.

3. Finish the shelf: After assembly, the shelf is ready for finishing, such as staining or painting. Clip 4 shows the staining process, where wood stain is applied to the shelves to enhance their appearance.

4. Install the shelves:  With the shelves built and finished, the next step is to mount them on the wall. Clip 1 clearly shows the installation process, including marking positions, drilling holes, inserting wall plugs, attaching wooden supports to the wall, and finally placing the shelf onto these supports.

\red{5. Showcase the final product: The final step is to display the completed shelves, often decorated with items. Clip 2 shows the finished floating shelves installed and decorated, with a person presenting them, indicating this is the final outcome.}

Arranging the clips in this logical order creates a step-by-step guide on how to build simple floating shelves.

\red{Correct order: Clip 5, Clip 3, Clip 4, Clip 1, Clip 2}

----------------------------------------------------------------------------------------------------------------------------------------------------------------------------------------------------------------------------------------------

\blue{\# Response from Gemini-1.5-Pro}

Certainly! Here's the correct order of clips to build the simple floating shelves, based on the logical progression of the project:

\red{Correct order: 2, 3, 4, 5, 1}

----------------------------------------------------------------------------------------------------------------------------------------------------------------------------------------------------------------------------------------------

\blue{\# Response from Qwen2.5-VL-Instruct$_\text{72B}$}

\red{Correct order: Clip 2, Clip 3, Clip 4, Clip 5, Clip 1}

\end{prompt}
}
\caption{
\textbf{Failure examples.} 
We observe that Gemini-2.0-Flash-Thinking generates a response with a detailed rationale explaining how it arrives at the final answer, unlike Gemini-1.5-Pro and Qwen2.5-VL-Instruct, which directly provide the final answer. Based on its detailed response, Gemini-2.0-Flash-Thinking correctly interprets most actions except for Clip 2 (highlighted in \red{red}). However, it entirely fails to arrange the identified events according to their causal dependencies. Additionally, both Gemini-1.5-Pro and Qwen2.5-VL-Instruct$_\text{72B}$ make the same mistake: they fail to recognize the causal link between steps $4$ and $5$, i.e., \textit{the shelves must be sanded before being painted}. 
The correct order is $2,3,5,4,1$.
}
\label{fig:qual_example}
\vspace{-0.15in}
\end{figure}

%% file: appx.tex
\setcounter{table}{0}
\setcounter{figure}{0}
\setcounter{equation}{0}
\renewcommand{\theequation}{S\arabic{equation}}
\renewcommand{\thetable}{S\arabic{table}}
\renewcommand\thefigure{S\arabic{figure}}

\section{Additional Results on \benchmark} \label{suppsec:benchmark_results}

We present the detailed results of LVLMs across varying numbers of causal steps in Table \ref{tab:step_wise_results} and their performance across different sub-categories of \benchmark in Table \ref{tab:detailed_results_class}.

\begin{table}[!h]
\centering
\fontsize{9pt}{10pt}\selectfont
\caption{
Performance across videos with varying numbers causal steps. LVLM performance drops sharply as the number of causal steps increases, highlighting challenges in handling complex video-based long-form causal reasoning tasks.}
\label{tab:step_wise_results}
\resizebox{0.8\linewidth}{!}{
\begin{tabular}{lcccccc}
\toprule
\multirow{2}{*}{\bf Models}
& \multicolumn{5}{c}{\bf Number of causal steps}
& \multirow{2}{*}{\bf Overall} \\
\cmidrule(lr){2-6}
& \bf 3 
& \bf 4
& \bf 5 
& \bf 6
& \bf 7
& 
\\
\midrule\midrule
InternVL2.5$_\text{38B}$ \cite{internvl25} & $15.9$ & $10.5$ & $10.0$ & $3.3$ & $0.0$ & $11.0$\\
InternVL2.5$_\text{78B}$ \cite{internvl25} & $18.2$ & $19.1$ & $8.8$ & $6.7$ & $4.2$ & $14.5$\\
Qwen2.5-VL-Instruct$_\text{72B}$ \cite{qwenvl25} & $40.5$ & $37.1$ & $16.2$ & $6.7$ & $4.2$ & $29.0$\\
GPT4o \cite{gpt4o} & $33.3$ & $40.0$ & $20.0$ & $13.3$ & $8.3$ & $29.0$\\
Gemini-1.5-Pro \cite{gemini15} & $60.3$ & $50.5$ & $36.2$ & $43.3$ & $20.8$ & $48.2$\\
Gemini-2.0-Flash-Thinking \cite{gemini15} & $64.8$ & $66.7$ & $46.2$ & $53.3$ & $29.2$ & $58.0$\\
\bottomrule
\end{tabular}
}
\vspace{-0.15in}
\end{table}

\begin{table}[!h]
\centering
\fontsize{9pt}{10pt}\selectfont
\caption{Detailed results across the sub-categories of \benchmark. 
}
\label{tab:detailed_results_class}
\setlength\tabcolsep{2pt}
\resizebox{\linewidth}{!}{
\begin{tabular}{lccccccccccccc}
\toprule
\bf Models  
& \rotatebox{90}{\specialcellleft{\bf Add Oil to\\\bf Your Car}}
& \rotatebox{90}{\specialcellleft{\bf Build Simple\\\bf Floating Shelves}}
& \rotatebox{90}{\specialcellleft{\bf Change a Tire}}
& \rotatebox{90}{\specialcellleft{\bf Grill Steak}}
& \rotatebox{90}{\specialcellleft{\bf Make Banana\\\bf Ice Cream}}
& \rotatebox{90}{\specialcellleft{\bf Make French\\\bf Strawberry Cake}}
& \rotatebox{90}{\specialcellleft{\bf Make French\\\bf Toast}}
& \rotatebox{90}{\specialcellleft{\bf Make Irish\\\bf Coffee}}
& \rotatebox{90}{\specialcellleft{\bf Make Jello\\\bf Shots}}
& \rotatebox{90}{\specialcellleft{\bf Make Lemonade}}
& \rotatebox{90}{\specialcellleft{\bf Make Meringue}}
& \rotatebox{90}{\specialcellleft{\bf Make Pancakes}}
& \rotatebox{90}{\specialcellleft{\bf Overall}}
\\
\midrule\midrule
InternVL2.5$_\text{38B}$ \cite{internvl25} & $6.5$ & $0.0$ & $8.0$ & $14.3$ & $16.7$ & $16.1$ & $0.0$ & $20.0$ & $16.7$ & $9.4$ & $7.7$ & $15.6$ 
& $11.0$\\
InternVL2.5$_\text{78B}$ \cite{internvl25} & $3.2$ & $6.9$ & $0.0$ & $11.4$ & $33.3$ & $19.4$ & $17.6$ & $10.0$ & $10.0$ & $15.6$ & $11.5$ & $31.2$ 
& $14.5$\\
Qwen2.5-VL-Instruct$_\text{72B}$ \cite{qwenvl25} & $16.1$ & $34.5$ & $4.0$ & $22.9$ & $43.3$ & $48.4$ & $20.6$ & $33.3$ & $26.7$ & $40.6$ & $19.2$ & $34.4$ 
& $29.0$\\
GPT4o \cite{gpt4o} & $22.6$ & $17.2$ & $4.0$ & $17.1$ & $70.0$ & $32.3$ & $20.6$ & $13.3$ & $43.3$ & $28.1$ & $38.5$ & $40.6$ 
& $29.0$\\
Gemini-1.5-Pro \cite{gemini15} & $12.9$ & $34.5$ & $0.0$ & $45.7$ & $93.3$ & $58.1$ & $47.1$ & $53.3$ & $53.3$ & $62.5$ & $34.6$ & $71.9$ 
& $48.2$\\
Gemini-2.0-Flash-Thinking \cite{gemini15} & $38.7$ & $55.2$ & $12.0$ & $65.7$ & $80.0$ & $61.3$ & $52.9$ & $65.5$ & $66.7$ & $59.4$ & $65.4$ & $65.6$ 
& $58.0$\\

\bottomrule
\end{tabular}
}

\end{table}

\section{Additional Details of \benchmark} \label{suppsec:benchmark_details}

\subsection{Details of procedures}
\label{sec:proc_details}

Table~\ref{tab:class_steps} lists the sub-categories included in \benchmark, along with the corresponding number of causal steps and number of samples for each sub-category. The number of causal steps varies across procedures to preserve meaningful causal relationships between intermediate steps and due to the natural diversity in how real-world procedural tasks are performed.

\begin{table}[!h]
    \centering
    \caption{Varying number of causal steps for different sub-categories of \benchmark.}
    \label{tab:class_steps}
    \begin{tabular}{lp{0.2\linewidth}l}
    \toprule
    \bf Sub-categories & \bf \# causal steps & \bf \# samples \\
    \midrule\midrule
    Add Oil to Your Car & 3, 4, 5 & 31 \\
    Build Simple Floating Shelves & 3, 4, 5 & 29 \\
    Change a Tire & 4, 5, 6, 7 & 25 \\
    Grill Steak & 3, 4, 5, 6, 7 & 35 \\
    Make Banana Ice Cream & 3, 4 & 30 \\
    Make French Strawberry Cake & 3, 4 & 31 \\
    Make French Toast & 5, 6, 7 & 34 \\
    Make Irish Coffee & 3 & 30 \\
    Make Jello Shots & 3 & 30 \\
    Make Lemonade & 3, 4, 5 & 32 \\
    Make Meringue & 4, 5 & 26 \\
    Make Pancakes & 3, 4, 5 & 32 \\
    \bottomrule
    \end{tabular}

\end{table}

\subsection{Details of human-level performance} \label{suppsec:human_eval}
\begin{figure}[!h]
    \centering
    \includegraphics[width=\linewidth]{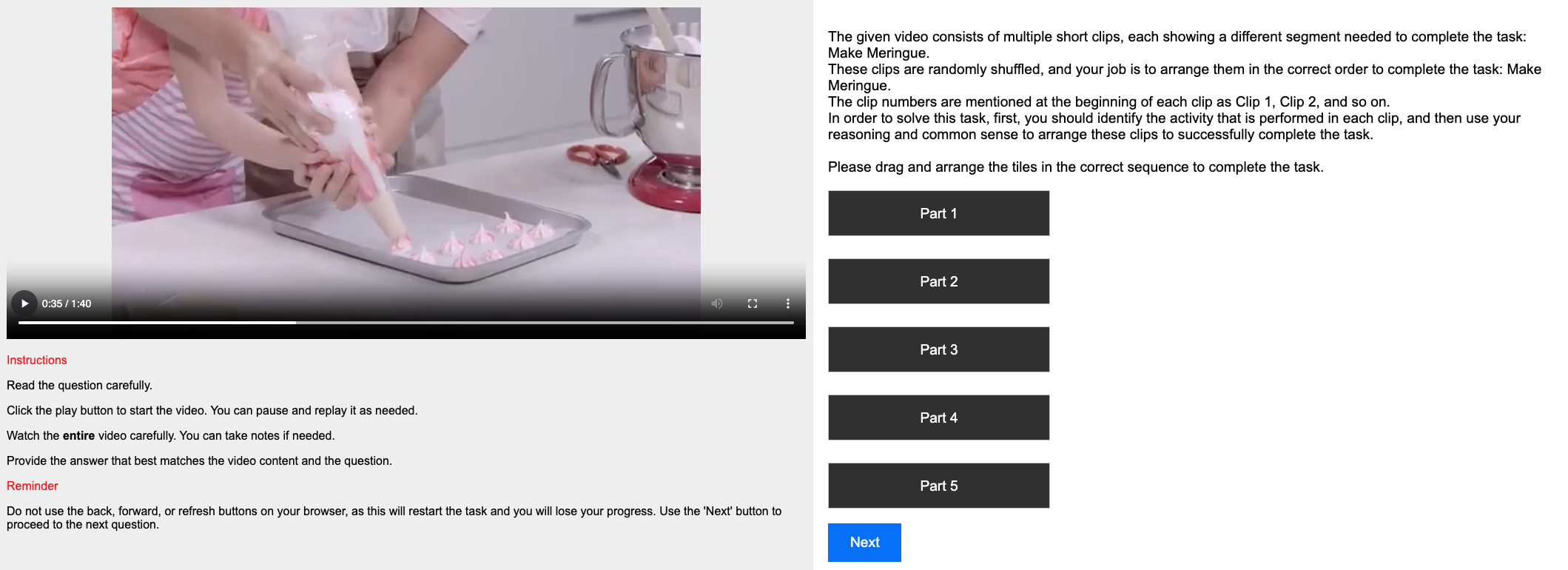}
    \caption{User interface for human evaluation. Questions are randomly shuffled to avoid any potential bias.}
    \label{fig:human_eval_setup}
\end{figure}

To obtain human-level performance on \benchmark, we recruit eight volunteers, who are undergraduate or graduate students. We collect their response on a representative subset of roughly $40\%$ of the videos and report the overall performance. An example of user interface is shown in Figure \ref{fig:human_eval_setup}.

\subsection{Licenses of existing assets used}
The videos used in constructing \benchmark are sourced from CrossTask \cite{crosstask} dataset, which is released under BSD-3-Clause license, available here: \url{https://github.com/DmZhukov/CrossTask?tab=BSD-3-Clause-1-ov-file}.

\subsection{Details of LVLMs}
The links to access LVLMs studied in this work are presented in Table \ref{tab:benchmarking_pp_url}.

\begin{table}[!h]
\centering
\fontsize{9pt}{10pt}\selectfont
\setlength\tabcolsep{1pt}
\caption{Details of LVLMs evaluated on \benchmark.}
\label{tab:benchmarking_pp_url}
    \begin{tabular}{ll}
    \toprule
    \bf Models & \bf Weights \\
    \midrule
    InternVL2.5$_\text{1B}$ \cite{internvl25} & \url{https://huggingface.co/OpenGVLab/InternVL2_5-1B} \\
    InternVL2.5$_\text{2B}$ \cite{internvl25} & \url{https://huggingface.co/OpenGVLab/InternVL2_5-2B} \\
    LongVU$_\text{3B}$ \cite{longvu} & \url{https://huggingface.co/Vision-CAIR/LongVU_Llama3_2_3B} \\
    InternVL2.5$_\text{4B}$ \cite{internvl25} & \url{https://huggingface.co/OpenGVLab/InternVL2_5-1B} \\
    VideoChat2$_\text{7B}$ \cite{videochat2} & \url{https://huggingface.co/OpenGVLab/VideoChat2_stage3_Mistral_7B} \\
    InternVL2.5$_\text{8B}$ \cite{internvl25} & \url{https://huggingface.co/OpenGVLab/InternVL2_5-1B} \\
    LLaVA-NeXT-Video$_\text{7B}$ \cite{llavanextvid} & \url{https://huggingface.co/LVLMs-lab/LLaVA-Video-7B-Qwen2} \\
    MiniCPM-o-V 2.6$_\text{7B}$ \cite{minicpm_v} & \url{https://huggingface.co/openbmb/MiniCPM-o-2_6} \\
    Qwen2.5-VL-Instruct$_\text{7B}$ \cite{qwenvl25} & \url{https://huggingface.co/Qwen/Qwen2.5-VL-7B-Instruct}  \\
    VideoLLaMA3$_\text{7B}$ \cite{videollama3} & \url{https://huggingface.co/DAMO-NLP-SG/VideoLLaMA3-7B} \\
    LongVILA$_\text{7B}$  \cite{longvila} & \url{https://huggingface.co/Efficient-Large-Model/qwen2-7b-longvila-256f} \\
    LongVU$_\text{7B}$ \cite{longvu} & \url{https://huggingface.co/Vision-CAIR/LongVU_Qwen2_7B} \\
    NVILA$_\text{15B}$ \cite{nvila} & \url{https://huggingface.co/Efficient-Large-Model/NVILA-15B} \\
    InternVL2.5$_\text{26B}$ \cite{internvl25} & \url{https://huggingface.co/OpenGVLab/InternVL2_5-26B} \\
    InternVL2.5$_\text{38B}$ \cite{internvl25} & \url{https://huggingface.co/OpenGVLab/InternVL2_5-38B} \\
    LLaVA-NeXT-Video$_\text{72B}$ \cite{llavanextvid} & \url{https://huggingface.co/LVLMs-lab/LLaVA-Video-72B-Qwen2} \\
    Qwen2.5-VL-Instruct$_\text{72B}$ \cite{qwenvl25} & \url{https://huggingface.co/Qwen/Qwen2.5-VL-72B-Instruct} \\
    InternVL2.5$_\text{78B}$ \cite{internvl25} & \url{https://huggingface.co/OpenGVLab/InternVL2_5-78B}  \\ %
    GPT4o \cite{gpt4o} & Accessed between Jan 2025 to Mar 2025 (gpt-4o-2024-11-20) \\
    Gemini-1.5-Pro \cite{gemini15} & Accessed between Jan 2025 to Mar 2025 (gemini-1.5-pro) \\
    Gemini-2.0-Flash-Thinking \cite{gemini15} & Accessed between Jan 2025 to Mar 2025 (gemini-2.0-flash-thinking-exp) \\
    \bottomrule
    \end{tabular}
\end{table}%

\section{Additional Details of RRD} \label{suppsec:rrd}

The complete instructions used in various RRD setups are mentioned in Figures \ref{fig:RRD_one_one} to \ref{fig:RRD_iter_iter}. 

\input{figures/rrd_inst}

\section{Broader Impact} 
Our proposed benchmark advances long-form causal reasoning in LVLMs, which is a critical yet underexplored area. Its novel task formulation avoids common linguistic shortcuts seen in multiple-choice and binary QA formats, allowing for a more reliable assessment of visual understanding. 
This design can generalize to other vision tasks to enable more accurate evaluation of LVLMs. We hope our benchmark will inspire further research on video-based reasoning and do not anticipate any new negative societal impacts resulting from this work.

%% file: figures/rrd_inst.tex
\begin{figure}

\begin{prompt2}

\blue{\# Instruction used in Video Recognition step.}

The video contains multiple short clips.

The clip numbers are mentioned at the beginning of each clip as Clip 1, Clip 2, and so on.

Watch each clip carefully, paying attention to its fine-grained actions and events. 

Note the unique events in each clip compared to the rest of the video.

Respond with a one sentence description indicating the key and fine-grained actions or events for each clip.

Please respond in this format:

Clip 1: <Write one sentence description>

Clip 2: <Write one sentence description>

...

Your response must not contain anything else.

\vspace{0.2in}
\blue{\# Instruction used in Causal Reasoning step.}

The following steps are needed to complete the task: \blue{\{name of the procedure\}}. 

However, these steps are randomly shuffled, and your job is to arrange them in the correct order to complete the task. 

Use your reasoning and common sense to arrange these steps to successfully complete the task.

\blue{\{clip descriptions\}}

The final output should be in this format:

Correct order: <mention the step numbers separated by a comma>
\end{prompt2}
\captionof{figure}{Instructions used in video recognition (all-at-once) and causal reasoning (all-at-once) setup.}
\label{fig:RRD_one_one}
\end{figure}

\begin{figure}
\begin{prompt2}

\blue{\# Instruction used in Video Recognition step.}

The video contains multiple short clips.

The clip numbers are mentioned at the beginning of each clip as Clip 1, Clip 2, and so on.

Watch Clip \blue{\{step\}} carefully, paying attention to its fine-grained actions and events. 

Note the unique events in Clip \blue{\{step\}} compared to the rest of the video.

Respond with a one sentence description indicating the key and fine-grained actions or events.

Your response must not contain anything else.

\vspace{0.2in}
\blue{\# Instruction used in Causal Reasoning step.}

The following steps are needed to complete the task: \blue{\{name of the procedure\}}. 

However, these steps are randomly shuffled, and your job is to arrange them in the correct order to complete the task. 

Use your reasoning and common sense to arrange these steps to successfully complete the task.

\blue{\{clip descriptions\}}

The final output should be in this format:

Correct order: <mention the step numbers separated by a comma>
\end{prompt2}
\captionof{figure}{Instructions used in video recognition (sequential) and causal reasoning (all-at-once) setup.}
\label{fig:RRD_iter_one}
\end{figure}

\begin{figure}
\begin{prompt2}

\blue{\# Instruction used in Video Recognition step.}

The video contains multiple short clips.

The clip numbers are mentioned at the beginning of each clip as Clip 1, Clip 2, and so on.

Watch each clip carefully, paying attention to its fine-grained actions and events. 

Note the unique events in each clip compared to the rest of the video.

Respond with a one sentence description indicating the key and fine-grained actions or events for each clip.

Please respond in this format:

Clip 1: <Write one sentence description>

Clip 2: <Write one sentence description>

...

Your response must not contain anything else.

\vspace{0.2in}
\blue{\# Instruction used in Causal Reasoning step.}

Here are two intermediate steps to achieving \blue{\{name of the procedure\}}:

Event A: \blue{\{description of one clip\}}

Event B: \blue{\{description of another clip\}}

Which event should occur first?

Pay attention to the causality of events.

Respond with A if Event A should happen first.

Respond with B if Event B should happen first.

Do not provide any other response.

\end{prompt2}
\captionof{figure}{Instructions used in video recognition (all-at-once) and causal reasoning (sequential) setup.}
\label{fig:RRD_one_iter}
\end{figure}

\begin{figure}
\begin{prompt2}

\blue{\# Instruction used in Video Recognition step.}

The video contains multiple short clips.

The clip numbers are mentioned at the beginning of each clip as Clip 1, Clip 2, and so on.

Watch Clip \blue{\{step\}} carefully, paying attention to its fine-grained actions and events. 

Note the unique events in Clip \blue{\{step\}} compared to the rest of the video.

Respond with a one sentence description indicating the key and fine-grained actions or events.

Your response must not contain anything else.

\vspace{0.2in}
\blue{\# Instruction used in Causal Reasoning step.}

Here are two intermediate steps to achieving \blue{\{name of the procedure\}}:

Event A: \blue{\{description of one clip\}}

Event B: \blue{\{description of another clip\}}

Which event should occur first?

Pay attention to the causality of events.

Respond with A if Event A should happen first.

Respond with B if Event B should happen first.

Do not provide any other response.

\end{prompt2}
\captionof{figure}{Instructions used in video recognition (sequential) and causal reasoning (sequential) setup.}
\label{fig:RRD_iter_iter}
\end{figure}